\documentclass[sn-apa]{sn-jnl}

\usepackage{graphicx}%
\usepackage{multirow}%
\usepackage{amsmath,amssymb,amsfonts}%
\usepackage{amsthm}%
\usepackage{mathrsfs}%
\usepackage[title]{appendix}%
\usepackage{xcolor}%
\usepackage{textcomp}%
\usepackage{manyfoot}%
\usepackage{booktabs}%
\usepackage{algorithm}%
\usepackage{algorithmicx}%
\usepackage{algpseudocode}%
\usepackage{listings}%
\usepackage{enumitem}
\usepackage{tabularx}
\usepackage{ragged2e}
\usepackage{array}
\usepackage{rotating}
\newcolumntype{L}[1]{>{\hsize=#1\hsize\RaggedRight} X}
\AtBeginDocument{}

\usepackage{fancyhdr} 
\fancypagestyle{firstpage}{
    \fancyhf{} 
    \fancyhead[L]{\textit{Published in Artificial Intelligence Review (2025). DOI: 10.1007/s10462-025-11341-4}} 
    \fancyfoot[C]{\thepage} 
    
}

\fancypagestyle{main}{
    \fancyhf{} 
    \fancyhead[CO]{\textsc{Macmillan-Scott \& Musolesi}} 
    \fancyhead[CE]{\textsc{(Ir)rationality in AI}}
    \fancyfoot[C]{\thepage}
    
}

\begin{document}

\title[\textbf{(Ir)rationality in AI}]{\textbf{(Ir)rationality in AI: State of the Art, Research Challenges and Open Questions}}

\author*[1]{\fnm{Olivia} \sur{Macmillan-Scott}}\email{olivia.macmillan-scott.16@ucl.ac.uk}

\author[1,2]{\fnm{Mirco} \sur{Musolesi}}

\affil[1]{\orgdiv{AI Centre, Department of Computer Science}, \orgname{University College London}, \orgaddress{\country{UK}}}

\affil[2]{\orgdiv{Department of Computer Science and Engineering}, \orgname{University of Bologna}, \orgaddress{\country{Italy}}}

\abstract{The concept of rationality is central to the field of artificial intelligence (AI). Whether we are seeking to simulate human reasoning, or trying to achieve bounded optimality, our goal is generally to make artificial agents as rational as possible. Despite the centrality of the concept within AI, there is no unified definition of what constitutes a rational agent. This article provides a survey of rationality and irrationality in AI, and sets out the open questions in this area. We consider how the understanding of rationality in other fields has influenced its conception within AI, in particular work in economics, philosophy and psychology. Focusing on the behaviour of artificial agents, we examine irrational behaviours that can prove to be optimal in certain scenarios. Some methods have been developed to deal with irrational agents, both in terms of identification and interaction, however work in this area remains limited. Methods that have up to now been developed for other purposes, namely adversarial scenarios, may be adapted to suit interactions with artificial agents. We further discuss the interplay between human and artificial agents, and the role that rationality plays within this interaction; many questions remain in this area, relating to potentially irrational behaviour of both humans and artificial agents. }




\maketitle

\thispagestyle{firstpage} 
\pagestyle{main}

\section{Introduction}\label{intro}
The way machines `think' and how close this can come to human reasoning has long been a topic of debate within artificial intelligence (AI) research. From Turing's proposal of a simple game to identify a machine that could think \citep{turing_1950}, to the 1956 Dartmouth Summer Research Project Proposal that sought to demonstrate that any aspect of learning or intelligence can be simulated by machines \shortcites{mccarthy_2006}\citep{mccarthy_2006}, the problem of understanding how far AI can push the boundaries of intelligence is inevitably intertwined with how machines reason. As such, the concept of rationality is central to the field of AI \citep{besold_2013}, whether we are referring to the rationality of artificial agents or of humans. Crucially, the goal is seldom to make an agent as rational as possible - some degree of irrationality is often desired, both in interactions amongst machines but also with humans. A more human-like artificial agent may at times elicit more trust \shortcites{waytz_2014}\citep{waytz_2014}, whereas in other scenarios simulating the fallibility of human reasoning can result in an artificial agent that appears untrustworthy. Despite the centrality of rationality to the field, there are still many open questions to be addressed.

Given the advent of large language models (LLMs) \shortcites{brown_2020}\citep{brown_2020}, as well as their increasing pervasiveness in our daily lives and their potential for societal impact \shortcites{eloundou_2023,kasneci_2023,li_moon_2023}\citep{eloundou_2023,kasneci_2023,li_moon_2023}, this problem has become even more fundamental. Generative AI (GenAI) systems raise new questions when it comes to rationality, particularly because the content they generate is of a form that humans ascribe meaning to. Models like LLMs have been shown to appear irrational \citep{binz_schulz_2023,macmillan-scott_2023,hagendorff_2023}, and this is not because these models are incorrectly achieving their set goals, but because we are evaluating them by metrics that differ from their training goals. While LLMs are not the only type of relevant artificial agent, they have highlighted how quickly AI can become integrated in our decision-making. If artificial agents are to be used in a decision-making capacity, whether this be medicine, diplomacy, or in our day-to-day, we must be sure that their output satisfies some clearly defined criteria and that we align the objective functions with these criteria. When it comes to human-AI interaction, if we do attempt to design more ‘human-like' machines, can we accept that they may at times mimic our own irrationality? These questions are crucial if we are to mitigate \textit{accidents}: unintended but potentially harmful consequences of AI, a concern that is key within AI safety \shortcites{amodei_2016,hendrycks_2022}\citep{amodei_2016,hendrycks_2022}.

Conversely, are there ways to leverage \textit{irrational} behaviour (both of humans and machines) in order to build more rational artificial agents? What are the best methods to identify and interact with irrational agents? As we develop increasingly complex and capable artificial agents, and as these become more integrated in everyday activities, an important question arises as to how the concept of rationality fits into their design. This may refer to assessing the rationality of the agent itself, but also to approaching interactions with other agents that may act in seemingly irrational ways, whether this is another artificial agent or a human. 

A number of differing definitions of rationality have been proposed within the field of AI, meaning that it is often unclear what is meant by a \textit{rational agent} \citep{hoek_wooldridge_2003}. Often, definitions have been adapted from our understanding of rationality in humans that has been developed in disciplines like economics, philosophy and psychology, rather than establishing a distinct definition of machine rationality. The notion of rationality plays a different role in different fields, and it is important to understand what conception is most suitable in each context \citep{besold_uckelman_2018}. \citet{knauff_2021_book} provide a comprehensive overview of the history and current understanding of rationality in philosophy, cognitive psychology and other disciplines, highlighting the varying notions that persist and advocating for more integrated interdisciplinary approaches. We cannot determine what it means to act or reason irrationally if we do not have a clear definition of rationality. However, these concepts are two sides of the same coin: we cannot define one without the other, and we will see that in certain scenarios, acting ‘irrationally' can in fact be the most rational approach.

Certain types of artificial agents already employ ‘irrational' behaviours in order to maximise reward, such as random actions taken as part of exploration in reinforcement learning \shortcites{ladosz_2022}\citep{ladosz_2022}, or profit non-maximising strategies that attempt to reduce potential loss \citep{ganzfried_2023}. As we will see, there are cases where irrational behaviour can lead to optimal outcomes. Crucial in determining when each type of behaviour is most suitable is understanding the environment that an agent is operating in, as well as potential agents one may interact with.

However, it is often not an easy task to identify what type of agents one is interacting with, or to establish their level of rationality. A number of opponent modelling methods have emerged that attempt to make predictions about an opponent, including their goals and actions \citep{albrecht_stone_2018}. For existing techniques, including policy reconstruction \shortcites{ganzfried_sandholm_2011,chakraborty_stone_2014,silver_2016,mealing_shapiro_2017}\citep{ganzfried_sandholm_2011,chakraborty_stone_2014,silver_2016,mealing_shapiro_2017} or classification \shortcites{abdul-rahman_2000,schadd_2007,iglesias_2010}\citep{abdul-rahman_2000,schadd_2007,iglesias_2010}, restrictive assumptions often need to be made that require some knowledge about the agent that is being modelled. Further developments in the identification of irrational agents will require domain-specific research, where employing combinations of existing methods presents a promising avenue for study.

 Similarly, once we have established that we are interacting with an irrational agent, what is the best strategy to follow? Current research has centred on interactions in adversarial domains \shortcites{bowling_veloso_2002,foerster_2018,letcher_2019,kim_2021,lu_2022}\citep{bowling_veloso_2002,foerster_2018,letcher_2019,kim_2021,lu_2022}, or with human opponents \shortcites{uprety_2018,mcelfresh_2021,chan_2021,azaria_2022}\citep{uprety_2018,mcelfresh_2021,chan_2021,azaria_2022}. It remains unclear how methods for adversarial environments can be adapted to employ in scenarios which may involve irrational agents. In particular, these methods may not be ideal when we want to achieve a cooperative outcome or are looking to obtain coordination between agents – one promising avenue of research that could be adapted to ensure cooperation are opponent shaping methods \citep{foerster_2018,letcher_2019,kim_2021,lu_2022}. Strategies must be adapted to the type of opponent, whether this be a human or artificial agent, therefore combining these methods with identification and classification techniques will likely be most effective. 
 
 While machines may be formulated to act as rationally as possible, humans have been shown to act and reason in irrational ways \citep{kahneman_tversky_1982}. We therefore need to design machines that are able to handle such scenarios and that do not assume perfect rationality of other agents. This is particularly important because we will see that perfect rationality is seldom the aim; instead, alternatives such as \textit{bounded rationality} or \textit{bounded optimality} are more realistic and often more desirable \citep{russell_1997}, so artificial agents are more likely to encounter boundedly rational opponents. 
 
In designing agents that achieve the optimal outcome, it may appear counter-intuitive to consider incorporating human cognitive biases. Nevertheless, we will explore how heuristics may be incorporated into decision-making in artificial agents to make them more efficient and improve their performance \shortcites{gulati_2022}\citep{gulati_2022}. As well as achieving greater capabilities, the use of cognitive biases inspired by human reasoning can increase the explainability of AI processes \shortcites{newell_shaw_simon_1959,simsek_2020}\citep{newell_shaw_simon_1959,simsek_2020}. However, building ‘irrational’ machines raises questions as to how human-AI interactions may be impacted. While the incorporation of human physical attributes has been shown to have a positive impact on human perception of artificial agents \shortcites{pizzi_2020}\citep{pizzi_2020}, questions remain as to whether the same effect is observed when concerning cognitive attributes, or even whether this is desirable in certain scenarios.

The aim of this article is to provide a comprehensive survey of rationality and irrationality within AI, as well as to set out the open questions in this area. Section \ref{sec2} introduces the varying definitions of rationality in different disciplines: the section begins by setting out how it is understood in AI and how this may be assessed, then surveys the definition of rationality in three disciplines that have most heavily influenced the conception within AI: economics, philosophy, and psychology. As we will see, the interdisciplinary nature of AI means that developments in computer science have been heavily influenced by notions of rationality in other fields. A comparison of all four disciplines is presented in Table \ref{disciplines}. Section \ref{sec3} details \textit{rationally irrational} behaviours, defined as behaviours that violate typical assumptions of rationality, but that can be shown to be rational under certain conditions (see Table \ref{types}). This section also includes a discussion of perceived irrationality in GenAI. Section \ref{sec4} then surveys existing methods for identifying and interacting with irrational agents; literature included in this section is summarised in Tables \ref{identification} and \ref{interaction}. Section \ref{sec5} focuses on the interplay between humans and AI: both how human irrationality can be incorporated into AI design in a beneficial way, as well as how the rationality of machines can have an impact on human-AI interactions. Section \ref{sec6} explores the new challenges brought about by GenAI, including tensions between what we train models to maximise for and how we evaluate their output. Finally, Section \ref{sec7} presents a number of open questions that remain to be addressed in this area – these constitute both fundamental theoretical questions, as well as potential avenues for research from a methodological standpoint.

\section{Defining \& Assessing Rationality} \label{sec2}
\subsection{Rationality in AI}\label{sec2.1}
\subsubsection{Defining Rationality}
The concept of a \textit{rational agent} has become an integral part of the AI discourse, although its precise meaning varies significantly between uses \citep{hoek_wooldridge_2003}. Whether considering an application in computer vision, LLMs, or reinforcement learning, the goal is invariably to arrive at a model that produces \textit{rational} decisions \citep{wooldridge_2000}. However, what is \textit{rational} can be interpreted a number of ways \citep{wheeler_2020}: for instance, it could mean making a decision closest to what a human would make, or achieving the optimal outcome, or even achieving a good outcome in the quickest time possible. \citet[p.273]{gresham_2015} define computational rationality as ``identifying decisions with highest expected utility, while taking into consideration the costs of computation." Within the field of AI, several differing approaches exist to the understanding of a rational agent. As we will see, developments within computer science have integrated conceptions of a rational agent from various fields \citep{zilberstein_2011}.

Whereas AI was first intended to replicate human intelligence \citep{brooks_1991}, it can be argued that pursuing this goal will not produce models that act in the most rational way \citep{wooldridge_2000}, therefore we first need to establish what the goal of AI is. A useful categorisation distinguishes the goal of artificial agents across two dimensions: reasoning vs. behaviour, and human vs. rational \citep{russell_norvig_2022}. The latter distinction exemplifies that often the ‘ideal’ reasoning or behaviour is not the one that most resembles human processes. In this definition, \textit{rational thinking} is defined in line with the \textit{laws of thought} approach, based in philosophy \citep{boole_2009,leech_2015}. This approach is grounded within the discipline of logic, however problems with defining a rational reasoning process often arise when it comes to the application, whether it be with expressing a problem formally or exhausting computational resources. On the other hand, an agent that exhibits \textit{rational behaviour} is one that acts so as to achieve the best outcome (or expected outcome); this result may be obtained by, but is not restricted to, correct inference (in line with rational reasoning). The distinction between rational reasoning and behaviour in AI is closely linked to the categorisation of theoretical (or epistemic) and practical rationality that is often used to study human rationality \citep{wedgwood_2021}. As we will see below, the study of epistemic rationality is more grounded within philosophy, whereas it is psychology that is most concerned with rational actions \citep{knauff_2021}.

It is often easier to implement and assess the rational behaviour aspect, as opposed to rational reasoning, as we can more easily observe behaviour. \citet{knauff_2021} frame this distinction using \citeauthor{marr_1982}'s \citeyearpar{marr_1982} levels of analysis, where his computational and algorithmic levels are referred to as output-oriented and process-oriented respectively. We are also often more concerned with output-oriented, computational mechanisms being rational rather than processes; that is to say, the focus is generally placed on evaluating the results of reasoning rather than the reasoning itself. Depending on the model architecture, pitfalls in behaviour may appear in different ways - an important one being the introduction of bias. The distinction between statistical learning and symbolic AI allows us to separate models that incorporate human biases through the data input, as opposed to those where an expression of ‘irrationality’ will come from the model architecture itself \citep{maruyama_2020} - both are subject to the incorporation of human bias, the variation arises in how the bias may be introduced. The former, statistical learning, can be defined as following a bottom-up approach. In this case, models generally learn from human data and inputs, and therefore the biases that appear are those that were introduced by the human input \citep{shin_2023}. Numerous examples exist of cases where the introduction of such biases has led to problematic results, most notably in computer vision \shortcites{buolamwini_2018,cook_2019,wilson_2019,noiret_2021,papakyriakopoulos_2023}\citep{buolamwini_2018,cook_2019,wilson_2019,noiret_2021,papakyriakopoulos_2023} and natural language processing (NLP) \shortcites{bolukbasi_2016,rozado_2020,dawkins_2021,hovy_2021,stanczak_2021}\citep{bolukbasi_2016,rozado_2020,dawkins_2021,hovy_2021,stanczak_2021}. On the other hand, biases in symbolic AI can be said to come from a more top-down approach. Given that these types of models or agents are generally designed to follow rules of logic and formal rationality, they do not generally exhibit the same biases as statistical learning \citep{maruyama_2020} – however, `irrational’ behaviour will present itself in different ways. In this case, it is the rules that underlie the model that are subject to human bias and may lead to irrational reasoning. There are increasingly examples of hybrid approaches that leverage a combination of the two, particularly advances in neuro-symbolic AI \shortcites{besold_2022}\citep{besold_2022}. 

Some of the discussion within statistical learning centres around removing the presence of human biases, or 
mitigating their presence within models and results. However, some have argued that there exist cases where
the use of heuristics similar to those used in human reasoning may make decision-making processes easier to explain \citep{newell_shaw_simon_1959,simsek_2020}. For instance, the use of tallying has been shown to be effective in many contexts, and to generalise to unseen problems \citep{dawes_corrigan_1974,einhorn_hogarth_1975,dawes_1979}. Similarly, the use of heuristics in Tetris can make AI more explainable \shortcites{simsek_2016}\citep{simsek_2016}. \citet{simsek_2020} argues that in linear environments, there are three conditions under which heuristics can 
yield accurate decisions: simple dominance \shortcites{hogarth_karelaia_2006}\citep{hogarth_karelaia_2006}, cumulative dominance \shortcites{kirkwood_sarin_1985,baucells_carrasco_hogarth_2008}\citep{kirkwood_sarin_1985,baucells_carrasco_hogarth_2008}, and noncompensatoriness \shortcites{martignon_hoffrage_2002,katsikopoulos_martignon_2006,simsek_2013}\citep{martignon_hoffrage_2002,katsikopoulos_martignon_2006,simsek_2013}. Developments in AI may not only render computational processes more explainable, but they may also reproduce processes in the human brain in such a way as to develop our understanding of how such processes occur. Until now, this research has predominantly focused 
on the use of deep neural networks \shortcites{cichy_2016,kubilius_2016,jozwik_2017,oconnell_chun_2018,bertolero_bassett_2020}\citep{cichy_2016,kubilius_2016,jozwik_2017,oconnell_chun_2018,bertolero_bassett_2020}, which have become in some cases a sort of ‘model organism’ \citep{scholte_2018,firestone_2020}. Through the development of the General Problem Solver (GPS), \citet{newell_simon_1961} presented a similar idea of using this program to aid in the construction of theories of human 
thinking.

The debate surrounding how we define rationality within AI also lends itself to the question of what we are trying to achieve. \citet{russell_1997,russell_2016} identifies four types of goals: perfect rationality, calculative rationality, bounded rationality, and bounded optimality. \citet{russell_norvig_2022} define them as follows: \textit{perfect rationality} occurs when an agent maximises expected utility with every action it takes - in most environments, this is not a realistic goal; \textit{calculative rationality} describes the case when an agent would eventually return what would have been the utility-maximising choice, but it may return it too late; \textit{bounded rationality}, as proposed by \citet{simon_1957,simon_1982}, follows the goal of \textit{satisficing} that will be discussed below; and \textit{bounded optimality} is more concerned with the decision-making process rather than the output - it refers to a case where agents take decisions as rationally as possible given their available computational resources, achieving an expected utility at least as high as any other agent with access to the same resources. While perfect rationality appears the ultimate goal and is theoretically possible, such as in the case of the Universal Turing Machine (UTM), the Halting Problem means we cannot know whether, when or how UTM will stop computing \citep{lee_2021}. 

There exist additional intrinsic limitations in accordance to Gödel's Incompleteness Theorems that relate to the Halting Problem; authors like \citet{lucas_1961} and \citet{penrose_1989}  have highlighted the consequences of the Gödel’s theorem as a limiting factor for machine rationality. Given that as a consequence of Gödel’s theorem there must always be a set of propositions that can neither be proven nor disproven, or a fact about the world that cannot be demonstrated as true, there will always exist problems that cannot be solved by machines. Therefore, it can be argued that striving for perfect rationality is a futile goal. In fact, the unsolvability of the Halting Problem has previously been used to prove Gödel's Incompleteness Theorem \shortcites{calude_2021,tourlakis_2022}\citep{calude_2021,tourlakis_2022}. 

Nevertheless, much of the debate in the field of AI centres around achieving perfect rationality. However, as we have seen there are constraints that, in many environments, make this an unrealistic goal. Instead of perfect rationality, \citet{simon_1957,simon_1982} introduced the aforementioned notion of \textit{bounded rationality} to explain how humans make decisions under uncertainty; he argued that individuals \textit{satisfice} rather than maximise: due to limits of cognitive ability or computation, time constraints, or incomplete knowledge, humans choose the first result that satisfies a chosen ‘aspiration level’ rather than obtaining the optimal solution \shortcites{gigerenzer_2020,schwarz_2022}\citep{gigerenzer_2020,schwarz_2022}. Further models of bounded rationality have emerged in the literature since \citep{rubinstein_1998,gigerenzer_selten_2002}. Surprisingly, the concept of bounded rationality has largely remained separate from the field of AI \citep{simsek_2020,lee_2021}. 

Arguably then, bounded optimality appears to offer the best framework: as opposed to calculative rationality, we can be sure of attaining a solution at the desired time, and we guarantee that this solution will have at least as high a utility as that achieved by a boundedly rational agent \citep{horvitz_1988,russell_1997}. Out of the four types of goals identified by \citet{russell_1997,russell_2016}, this comes closest to the above definition of computational rationality given by \shortcites{gresham_2015}\citet{gresham_2015}. There may even be cases where bounded optimality cannot be achieved, therefore \citet{russell_s_p_1993} define a weaker property, asymptotic bounded optimality, as a more robust and tractable alternative. \citet{zilberstein_2011}, who instead classifies bounded optimality as one amongst other approaches to bounded rationality, concurs that it is difficult to achieve in practice. Instead, he proposes \textit{metareasoning} as a more promising approach - more specifically, optimal metareasoning. Inspired by a component of human decision-making, metareasoning can be defined as a way of allocating resources: it is a ``mechanism to make certain runtime decisions by reasoning about the problem solving – or object-level – reasoning process" \citep[p.29]{zilberstein_2011}.

It is clear that there is no comprehensive definition of rationality that serves every purpose. Although perfect rationality appears to be a sensible aim, we have seen there are many limitations to achieving this in practice (namely the Halting Problem). A similar issue arises with bounded optimality, which is often difficult to attain. Bounded rationality is a more realistic goal, although in implementing this we must accept that we will not always achieve the optimal outcome.

\subsubsection{Assessing Rationality}
When assessing whether an agent has attained the level of rationality we are seeking, a crucial question lies in how we measure its performance. Taking inspiration from developmental and comparative psychology, the performance vs. competence debate can be applied to artificial agents \citep{firestone_2020,lampinen_2023}. The debate highlights the distinction between \textit{competence}, defined as a system’s internal knowledge and capabilities, and \textit{performance}, which relates to demonstrations of this knowledge. A \textit{species-fair} comparison between humans and machines requires ensuring that tests account for both human and machine constraints, and implement task alignment. If tasks are not defined with this notion in mind, poor performance may be taken as a sign of low competence, when in fact it arises from tests that are not well suited to the constraints of either the human or machine being evaluated.

The type of test we require to assess an agent’s rationality also depends on what aspect we are interested in. Returning to the distinction between reasoning and behaviour, our method of assessment will vary for each of these. Often, the characteristic of interest is rational behaviour, as we are interested in the result. To assess whether behaviour is rational, we can apply the rational agent approach: \textit{an agent that acts so as to achieve the best outcome, or best expected outcome under uncertainty, would be deemed a rational agent}. This is in contrast to a test for rational thinking, which would ascertain whether the agent reasons in accordance with the \textit{laws of thought}. Recent work has also questioned classical definitions of rationality when it comes to conditional logic \citep{eichhorn_2018}, defining a set of non-monotonic inference patterns and assessing them based on the plausibility semantics of Ordinal Conditional Functions (OCF) \citep{spohn_1988}. In so doing, the authors define a new way of evaluating what may be deemed rational according to plausible reasoning standards, with a requirement of consistency, applicable to both human reasoning and artificial agents. While here we are concerned with rational reasoning and behaviour, other types of assessment would be necessary to determine whether an agent is behaving or reasoning in a human way; the Turing test \citep{turing_1950} is seen as an example of this type of assessment, even though it has been criticised for its inherent limitations \citep{hernandez-orallo_2000}. 

Assessing the rationality of GenAI systems poses a new set of challenges. Even more clarity about what and how we are evaluating is needed, as these systems encompass often conflicting types of rationality. Taking LLMs as an example, these models are trained with the goal of predicting the most likely next token. A model may exhibit high performance in this sense, but produce an output that appears irrational to a human interpreter. Because the output of LLMs are in a form that humans ascribe meaning to (language), we evaluate their responses as we would human speech. Therefore, there is a tension between the goal of the model (next-token prediction) and the way we evaluate its output under a human lens. We refer to this problem as the \textit{GenAI Evaluation Paradox}. There is therefore a need to align the goals considered in training and evaluation methodologies. When evaluating LLMs, we often look at the correctness of the responses or the coherence of reasoning, but this is not what LLMs have been trained to maximise. As we will see below, alignment and safety fine-tuning is generally at odds with the model's attempt to minimise next-token prediction loss. This argument also holds for other types of outputs produced by GenAI systems, like images or audio. The question of GenAI and its implications of rationality will be further explored in Section \ref{sec3.5} and Section \ref{sec6}.

\subsection{Rationality in Other Disciplines}\label{sec2.2} 
\subsubsection{Economics}
The concept of rationality is a central assumption in certain areas of economics. In its definition, particularly within normative decision theory, rationality has become analogous to consistency \citep{hammond_1997,schiliro_2012} – as such, close parallels can be drawn to the definition of rationality in philosophy, as will be seen below. A rational action or decision is therefore one that maximises utility, or expected utility where there is incomplete information. While in other disciplines the focus generally lies within rational behaviour or reasoning, within economics, it is extended to questions of preferences, beliefs, expectations and knowledge \citep{hammond_1997}. For instance, the consistency of ordinal preferences is often emphasised, where what matters is the order of these preferences (see \citet{hicks_allen_1934}).

Economic theory establishes a quantifiable way of assessing rationality based on measures of utility. However, real values of utility can seldom be measured in an objective way. As a response to this problem, \citet{samuelson_1938} developed the theory of revealed preference, focusing on observed demand behaviour to infer underlying preferences. Ultimately, looking at demand behaviour to develop an understanding of the underlying preferences can be assimilated to the reasoning vs. behaviour distinction that was discussed above: where we cannot assess the rationality of reasoning, as we do not observe this process, we can instead evaluate the rationality of observed actions. The theory of revealed preference was later developed, building on Samuelson’s work, to further establish how to infer preferences from demand functions \citep{houthakker_1950,arrow_1959,uzawa_1960,afriat_1967}. 

Although it can be argued that economic theory ultimately seeks to study and predict human behaviour \citep{sugden_1991}, its concept of rationality is, unlike human behaviour, perfect, logical and deductive \citep{arthur_1994}. \citet{gintis_2000} highlights this notion, arguing that following the traditional economics definition of rationality renders humans `hopelessly irrational.' The disconnect between the theoretical and empirical evidence of behaviour \shortcites{tsesos_2016}\citep{tsesos_2016}, grounded in the fact that humans cannot reason beyond given levels of complexity, is addressed by the aforementioned concept of bounded rationality. Thaler, who collaborated with and built on the work of Kahneman and Tversky (see for example \shortcites{thaler_1980,tversky_thaler_1990,kahneman_knetsch_thaler_1991}\citet{thaler_1980,tversky_thaler_1990,kahneman_knetsch_thaler_1991}), incorporated findings from psychology into our understanding of economics. The field of behavioural economics was pioneered by Thaler, and there is now extensive research within economics surrounding cognitive biases and heuristics \shortcites{slovic_2002,grandori_2010,zindel_2014,enke_2023}\citep{slovic_2002,grandori_2010,zindel_2014,enke_2023}.

So far we have considered expected utility theory as a descriptive or predictive theory that sets out to either explain how humans make decisions or predict the choices they will make. However, another perspective offers a normative view of expected utility theory \citep{cave_2005}, seeing it instead as a way of studying how humans \textit{should} rather than \textit{do} make decisions \citep{briggs_2023}. From this perspective, the irrationality of observed behaviour in our decision-making does not contradict the theory; rather, expected utility theory may be more in line with the aforementioned perfect rationality approach. If we consider applications of expected utility theory to machine learning, in particular reinforcement learning, the interpretation as a normative theory becomes apparent \citep{parkes_wellman_2015}: expected utility is used to decide what action an agent should take next, and in this way find an optimal policy to follow (see for example \citet{charpentier_2020}). Learning algorithms have further incorporated data from human psychological experiments to model choice behaviour and human decision-making \citep{brian_arthur_1993}.

\subsubsection{Philosophy}
When defining rationality within artificial agents, Aristotle’s ‘laws of thought’ approach has been employed as a way of defining and assessing rational thinking in machines \citep{russell_norvig_2022}. For the ancient Greek philosophers, however, rationality was a uniquely human trait that sets us apart from other animals \citep{nozick_1993}. The conception of rationality within philosophy can similarly be attributed to Aristotle, whose classification of the virtues led to the distinction between \textit{epistemic} (or theoretical) rationality and \textit{practical} rationality \citep{rysiew_2008,okasha_2016,wedgwood_2021}. This categorisation of the two types of rationality, relating to beliefs and decisions respectively, is closely linked to the separation between reasoning and behaviour discussed above. As understood in the epistemic sense, rationality can often be equated to beliefs that are justified, and is closely linked to truth \citep{rysiew_2008}; crucially, rational belief must be supported by reasons and be generated through a reliable process \citep{nozick_1993}. Conversely, practical rationality concerns actions and desires \citep{battaly_slote_2015}, and is generally more the concern of psychological study. \citet{foley_1987} points to issues with this distinction, and argues that inconsistent beliefs may still be rational from an epistemic standpoint.

The current debate within epistemology is centred between internalist and externalist perspectives. The former’s origins are sometimes traced back to Hume \citep{meeker_2001}, although some have argued that this attributed view points to an inconsistency in his moral theory \citep{brown_1988,coleman_1992}. In the internalist view, reason lies at the core of rational belief, and is as such internal to the person. A belief may be false, however if there is justification for this belief, it is not necessarily irrational to maintain said belief \citep{audi_2002}. In contrast, externalists ascertain that rational thinking and the justification of a belief is not only a result of internal processes, but is also affected by external factors \citep{farkas_2003}. 

Other definitions of rationality within philosophy centre on reasoning with a logical grounding. \citet{stein_1996} developed a \textit{Standard Picture} of rationality, where the emphasis is on reasoning according to principles based on logic, probability theory and so forth. In trying to model how the mind works, \citet{sloman_1993} used computers as a way to represent our own rationality, equating the mind to a \textit{control system}. A similar framework to the \textit{Standard Picture} is the ‘classical’ conception discussed by \shortcites{chase_h_g_1998}\citet{chase_h_g_1998}, who argue for the ecological rationality of heuristics used by humans and organisms more generally. \citet{evnine_2001} sets out a theory of the \textit{Universality of Logic}, which contends that rational creatures must possess certain logical abilities. A more Kantian perspective has been suggested, defending the \textit{logic-oriented conception of human rationality}, claiming that rational animals are intrinsically logical animals \citep{hanna_2006}.

However, similar to economics, the concept of rationality within philosophy has also been impacted by empirical studies in psychology. The debate around cognitive biases and limits in human reasoning have given rise to questions surrounding how rationality is defined \citep{rysiew_2008}: if humans do not follow the formal norms of rationality, does this render them irrational, or do we need to redefine our conception of rationality? Some see rationality as an evolutionary adaptation, and in this conception, humans cannot be deemed irrational \citep{dennett_1981,obrien_1993}. In this line of argument, an explanation can be found for any ‘irrationalities’ displayed by humans. Others argue that our definition of rationality should be based on human intuition \citep{cohen_1981}, and that any mistakes are due to \textit{performance} rather than \textit{competence} errors. If we are to define rationality in accordance to the way humans reason, how would this then impact the conception of machine rationality? It may be that we need a definition of rationality exclusive to humans, and another for machines. 

Others emphasise the distinction between \textit{normative} and \textit{descriptive} rationality, mirroring the discussion within economics. In this view, philosophy would be more concerned with normative rationality, whereas psychology centres on descriptive rationality \citep{knauff_2021}. The two are intrinsically linked, as, for instance, rational action will typically depend on coherent beliefs. For this reason, \citet{knauff_2021} argue for a more integrated study of normative and descriptive approaches, advocating for an interdisciplinary research agenda. 
    
\subsubsection{Psychology}

With regards to defining rationality in psychology, much of the focus in the literature has been on understanding how humans make decisions, placing the emphasis on practical and descriptive rationality. Significant study in psychology has been on human cognitive biases and limitations in human reasoning, and whether this deems humans irrational, or whether we need to redefine what rationality means in relation to human behaviour. This debate has formed two major camps, denoted by \citet{sturm_2012} as the `heuristics and biases’ approach versus the ‘bounded rationality’ approach, and has been referred to as the ‘rationality wars’ in the psychology of human reasoning \citep{samuels_s_b_2002}. The ‘heuristics and biases’ literature, where the most defining work was carried out by Tversky and Kahneman, contends that humans often do not reason according to rules of logic and probability (see for example \citet{wason_1968,tversky_kahneman_1971,kahneman_tversky_1982,tversky_kahneman_1986,kahneman_knetsch_thaler_1991}; for collections of such works, see \shortcites{nisbett_ross_1980,gilovich_griffin_kahneman_2002}\citet{nisbett_ross_1980,gilovich_griffin_kahneman_2002}). While research in this area often emphasised the irrationality in human reasoning, others have proposed a novel way to evaluate rational reasoning in conditional logic, arguing that even though some inference patterns displayed by human reasoners are not valid according to classical logic, they are nevertheless plausible and should thus be considered rational \citep{eichhorn_2018}. The authors emphasise consistency in reasoning as necessary for rationality, rather than classical logic.

In contrast to the ‘heuristics and biases’ literature, the `bounded rationality’ scholarship argues that human beings are not irrational in the way they reason and make decisions – the way humans reason is rational given computational limitations, as well as limits in knowledge and attention \citep{gigerenzer_1993,gigerenzer_goldstein_1996}. \shortcites{lewis_2014}\citet{lewis_2014} develop a framework of \textit{computational rationality} that looks at bounds on decision-making mechanisms within the brain and attempts to unify explanations of behaviour and mechanisms within psychology, tying in with the utility maximisation approach from economics. Whereas some have argued that the two perspectives can be reconciled, as it is a question of how much and when we follow certain norms \citep{samuels_s_b_2002}, others do maintain that the arguments are substantively different, although not wholly incompatible \citep{sturm_2012}.

In their work on heuristics, \citet{kahneman_tversky_1982} identified a number of cognitive biases present in human reasoning that contradict the laws of thought and logic, such as the Linda Problem (conjunction fallacy) \citep{tversky_kahneman_1983} or the Gambler’s Fallacy \citep{tversky_kahneman_1971}. The results have been replicated empirically numerous times, showing under what conditions humans appear to violate these norms, as well as the conditions that reduce the presence of these biases \shortcites{nisbett_borgida_1975,clotfelter_cook_1993,donovan_epstein_1997,tentori_b_o_2004,barron_leider_2010}\citep{nisbett_borgida_1975,clotfelter_cook_1993,donovan_epstein_1997,tentori_b_o_2004,barron_leider_2010} (in recent years, similar experiments have been done with LLMs instead of humans \citep{macmillan-scott_2023,binz_schulz_2023,hagendorff_2023}). Although these biases seem to illustrate irrationality within human reasoning, some argue that these heuristics and biases are often useful when applied to the right domains – generally domains similar to ones in which they emerged – and can be remarkably effective \citep{kahneman_tversky_1982,gigerenzer_goldstein_1996}.

Given the existence of these heuristics that appear to show that human reasoning often contradicts formal norms of rationality, recent studies in psychology and neuroscience have argued that there exist two types of reasoning processes within human cognition. System-1 is associated with fast, automatic, and largely unconscious processes, and is thought to be a product of evolution; in contrast, System-2 is slower, rule-based, and subject to deliberate conscious control, it is more plastic and is shaped both by culture and education \citep{evans_over_1996,sloman_1996,stanovich_1999,rysiew_2008}. The identification of these two processes has led to a discussion around how we define rationality within human reasoning, and the potential of splitting our understanding of rationality in such a way that it mirrors the two cognitive systems \citep{evans_over_1996,sloman_1996,stanovich_1999,saunders_over_2009}. The debate had expanded into the AI literature; \citet{bengio_2019} assimilates current deep learning capabilities to System-1, and discusses how recent developments are paving the way for deep learning architectures with capabilities that include System-2 tasks, which require conscious reasoning.

Considering the differing conceptions of what rationality means within psychology, it is unsurprising that there is no unified test for rational thinking. One important distinction that is often made is that between rationality and intelligence \citep{sutherland_1992}. On one hand, within the field of AI, \citet{russell_1997} sets out potential definitions of intelligence as the types of rationality or optimality covered above, in so doing establishing machine intelligence as rationality. Similarly, \shortcites{gresham_2015}\citet{gresham_2015} discuss a computational rationality framework as a way to study intelligence (both in brains and machines), again equating the two notions. On the other hand, within psychology there is a clear distinction. \shortcites{stanovich_west_toplak_2011}\citet{stanovich_west_toplak_2011} see rationality as a much broader notion than intelligence that encompasses more cognitive skills, particularly if we define intelligence by what is measured in commonly used intelligence tests. They see rationality as a superordinate concept to intelligence given that it encompasses ``both the reflective mind and the algorithmic mind" \shortcites{stanovich_west_toplak_2011}\citep[p.814]{stanovich_west_toplak_2011}. In the same way that we have developed a way to measure intelligence through the use of the IQ test, \citet{stanovich_2016} proposes an assessment of rational thinking called the CART (Comprehensive Assessment of Rational Thinking). Other types of assessment are used empirically for particular aspects of rationality, such as the use of dominance tests to study choice behaviour \shortcites{tervonen_al_2018}\citep{tervonen_al_2018}. Dominance tests are often used in discrete choice experiments (DCEs) \shortcites{ryan_2001,McIntosh_2002}\citep{ryan_2001,McIntosh_2002}, and have therefore been used predominantly within economics – as such, the definition of rationality in this type of assessment must be in line with axioms of rational choice, including completeness, transitivity, and monotonicity \shortcites{mascolell_whinston_green_1995}\citep{mascolell_whinston_green_1995}.

A similar debate around how to define rationality exists in neuroscience, where more emphasis has been placed on pathology and how this relates to rational reasoning. A crucial question concerns whether pathology indicates an impairment on rationality, or whether mental illness does not necessarily signify irrationality. While some psychopathology patients may exhibit irrational behaviour, many hold the view that irrationality is not sufficient or necessary for mental illness \citep{bortolotti_2013,cardella_2020}: insanity and irrationality are not intrinsically linked. If we take the goal of the brain as minimising surprise or uncertainty \citep{friston_2010}, both healthy and diseased brains can be considered rational in so long as they pursue this goal \citep{fiorillo_2017}. Conversely, research has shown that lesions to the orbitofrontal cortex may hinder the anticipation of negative emotional consequences \shortcites{kirman_2010}\citep{kirman_2010}. Accounting for these consequences can lead to better informed decision-making and choice behaviour, therefore some pathologies may indeed affect rational reasoning. Perhaps counterintuitively, emotions influence and can be considered essential to rational thinking \citep{martins_2011}. Therefore, it is neither pathology nor emotions that cause the biases in judgement that are observed in our decision-making.

\vspace{7pt}

\noindent Table \ref{disciplines} presents a comparative analysis of rationality and irrationality across the domains of AI, economics, philosophy, and psychology.

\begin{table}
\caption{Rationality and irrationality in different disciplines.}\label{disciplines}
\begin{tabularx}{\textwidth}{L{0.2} L{0.2} L{0.2} L{0.2} L{0.2}} 
\toprule%
  \multicolumn{1}{c}{\textbf{}} & \multicolumn{1}{c}{\textbf{AI}} & \multicolumn{1}{c}{\textbf{Economics}} & \multicolumn{1}{c}{\textbf{Philosophy}} & \multicolumn{1}{c}{\textbf{Psychology}} \\
\midrule
 How is rationality determined? & Optimal outcome & Utility maximisation & Logical reasoning & Contested (Rationality wars) \\
 \midrule
 How does rationality relate to preferences? & Rationality requires consistent preferences & Rationality requires consistent preferences & Rational preferences are arrived at through justified reasoning & Rationality requires consistent preferences \\
 \midrule
 What is meant by bounded rationality? & Limited computation/time & Finite recursion & Human computational limitations, limits on knowledge and attention & Human computational limitations, limits on knowledge and attention \\
 \midrule
 What is considered irrational? & Taking an action that does not maximise expected utility & Taking an action that does not maximise expected utility, inconsistent preferences & Unjustified, inconsistent beliefs or actions &  Human cognitive biases and heuristics (although contested) \\
 \midrule
 Rational Reasoning & Correct inference  & Consistent preferences & Justified beliefs (epistemic rationality) & Adherence to rules of probability and logic \\ 
 \midrule
 Rational Behaviour & Rational agent - acts so as to achieve best expected outcome & Taking action that maximises expected utility & Justified actions / desires (practical rationality) & Acting according to rational reasoning \\
\botrule
\end{tabularx}
\end{table}

\section{Rational Irrationality}\label{sec3}

Having established that theories and definitions of rationality vary significantly across and within disciplines, it is a challenge to then establish what constitutes an irrational agent. Agents may deviate from rational reasoning or behaviour in different ways, and may be considered rational or not depending on the definition we are working off. As we have seen, it may be that we need to develop differing definitions for rationality in humans and machines, or that we need to reevaluate our understanding of rationality altogether. \citet{eichhorn_2018} demonstrate that some human inferences labelled as irrational can in fact reflect rational strategies when judged within a conditional plausibility framework rather than by means of classical logic. In this section, we will be adhering to the aforementioned notion of perfect rationality \citep{russell_1997}; that is to say, we define a rational agent as one that invariably takes actions that maximise its expected utility, given the available information. 

In this section, we will discuss types of irrationality – these are deviations from perfect rationality in different ways. However, as we will see below, the term `irrational' must be taken with caution. Each type of irrationality that we discuss below can in fact constitute the optimal behaviour or reasoning in the right scenario. For example, we discuss bounded rationality in this section: bounded rationality is generally considered as rational behaviour, however we include it here as it deviates from perfect rationality. In fact, \citet{gigerenzer_2001} distinguished between nonrational theories of decision-making and theories of irrational decision-making; this distinction emphasises that non-rational theories may not necessarily lead to suboptimal outcomes in the real world. The categorisation included here is not an exhaustive list, but presents the cases most studied in the literature and are all behaviours that commonly occur in practice. The cases presented here illustrate the notion that irrational behaviour can sometimes emerge as preferable or even optimal, and are summarised in Table \ref{types}. 

The cases where irrational behaviour can, in some scenarios, be optimal are distinguished from perceived irrationality. Perceived irrationality is particularly relevant when it comes to GenAI, and as such is briefly discussed below. We return to a fuller consideration of GenAI in Section \ref{sec6}. Whereas as an agent may be perceived to be irrational when for instance their goals are unknown, irrational behaviour is denoted as that where no matter the goal, the behaviour is suboptimal \citep{masters_sardina_2021}. This may arise particularly in situations involving limited information – agents may not be aware of other agents’ utility functions, and may have different reward structures in such a way that behaviour appears to be irrational but could in fact be optimal for another agent. In game theoretical frameworks, an agent may appear irrational when their model of the game differs from our own \citep{ganzfried_2023}. Behaviour being perceived as irrational has been studied in the domain of human-computer interaction, where perceived irrationality may arise from the discrepancy between an individual’s expectations and their perception of a smart system’s actions \shortcites{abadie_2019}\citep{abadie_2019}. As we will see in more detail in Section \ref{sec5.2}, the question of how rational a system should be when interacting with humans remains an open question. 

\subsection{Bounded Rationality}
Bounded rationality is one of the most interesting characterisations of possible types of rationality. Proposed by Herbert Simon, bounded rationality seeks to explain how humans make decisions under uncertainty, and puts forward the idea of \textit{satisficing} rather than maximising \citep{simon_1957,simon_1982}. As mentioned above, the principle behind \textit{satisficing} is that due to given constraints, humans choose the first result that satisfies a chosen ‘aspiration level’ rather than obtaining the optimal solution \shortcites{gigerenzer_2020,schwarz_2022}\citep{gigerenzer_2020,schwarz_2022}. Although it was developed to better understand human decision-making, bounded rationality can also be applied to artificial agents and their behaviour under uncertainty or computational constraints \citep{russell_1997}.

However, until now, research on artificial agents has not greatly overlapped with the study of bounded rationality \citep{simsek_2020,lee_2021}, therefore open questions remain on the behaviour and interaction of agents with limited computation. The constraints that exist in human reasoning differ from those of artificial agents, as does the threshold for what can be deemed as a \textit{satisficing} solution. With regards to artificial agents, \shortcites{wen_2020}\citet{wen_2020} study this problem in the context of recursive reasoning in multi-agent interactions. They introduce the Generalized Recursive Reasoning (GR2) framework to model the dynamics in interactions of agents with differing levels of rationality – in this case, the levels of rationality are interpreted as levels of recursive reasoning. The motivation for this research is that, in interactions with irrational (non-optimal) agents, the effectiveness of existing MARL models significantly decreases \shortcites{shoham_2003}\citep{shoham_2003}. In their model, an agent with a given level of recursive reasoning takes the best response to all possible lower-level agents. In this sense, the boundedly rational agent is taking the best response to other agents whose rationality is bounded to a higher degree, meaning that each agent is determining the best response when interacting with agents that have a lower level of recursive reasoning and therefore whose ratiomality is more bounded. While this generates valuable results, questions remain surrounding the behaviour of agents when interacting with opponents whose rationality is less bounded – in this example, looking at environments with agents operating at higher levels of recursion. 

\subsection{Random Behaviour} 
Random behaviour is behaviour that appears to have no rationale or clear motivation, and as such is interpreted as irrational due to the lack of an identifiable goal. It can be distinguished from other types of irrationality that, although suboptimal, still adhere to a recognisable strategy. However, random actions are often crucial to the behaviour of artificial agents, particularly in situations of bounded rationality or limited resources. When interacting with human participants, artificial agents that incorporate some degree of randomness have also been shown to improve collective performance \citep{shirado_2017}. \citet{icard_2021} highlights two compelling rationales for randomisation: costly computation, where establishing the best course of action requires resources, and finite memory. He concludes that while there may be Bayesian justifications against randomising behaviour, there will arguably always be cases where a \textit{satisficing} solution, following Simon’s definition, requires some randomising. 

Within the field of reinforcement learning, random exploration has emerged as the most common technique for exploration \citep{ladosz_2022} – in particular, $\varepsilon$-greedy exploration, which maintains a balance between \textit{exploration} and \textit{exploitation} \shortcites{kaelbling_1996,bloembergen_2015}\citep[e.g.,][]{kaelbling_1996,bloembergen_2015}. Whereas exploration employs random behaviour to look for potentially higher rewards, exploitation sticks to known high rewards. Therefore, random behaviour does not necessarily equate to irrational behaviour – where there is a longer-term motivation, taking random actions may in fact allow an agent to arrive at the optimal strategy \shortcites{still_precup_2012}\citep{still_precup_2012}. The existence of this long-term motivation can constitute the determinant of whether this behaviour can be attributed rationality: random behaviour can in fact lead to optimal decision-making in the long run. 

Within research on reinforcement learning, and particularly deep reinforcement learning, there is evidence of the advantages of random exploration \shortcites{fernandez_2006,mnih_2015,polvara_2018,yu_2020}\citep{fernandez_2006,mnih_2015,polvara_2018,yu_2020}. However, there are also downsides to this type of behaviour, namely its inefficiency as it often revisits the same states. \citet{ladosz_2022} consider three approaches to address this inefficiency: reduced states/actions for exploration methods, exploration parameters methods, and network parameter noise methods. Each of these presents its own improvements and disadvantages; for instance, exploration parameter methods are particularly useful in achieving a good balance between exploration and exploitation, but does not fully solve the question of inefficient exploration of previously seen states. Nevertheless, these methods based on random exploration exemplify that such a behaviour is not inherently irrational.

Similarly, randomisation holds an important role within stochastic games. In game theory, the Nash Equilibrium is often used to signify the optimal strategy given available information. While these solutions are sometimes constituted by pure strategy Nash Equilibria, often there also exists a mixed solution \shortcites{daskalakis_2009,bloembergen_2015}\citep{daskalakis_2009,bloembergen_2015}. Mixed solutions in game theory involve randomisation over a set of strategies with a given probability distribution \citep{leyton-brown_2008}. Therefore, in such cases, randomness exists as part of the `ideal' rational behaviour \citep{icard_2021}. As with reinforcement learning, it is then the use of randomness with a broader motivation that renders this type of behaviour ultimately rational in these scenarios. The motivation behind random actions may not always be known to other agents in the interaction, and as such behaviour that appears random may only be perceived irrationality or may be deceptive. 

\subsection{Profit Non-maximising}
Revisiting the definition of a perfectly rational agent as one that takes actions to maximise utility at every instance, an agent that, given the same information, does not take these actions seems to provide a clear indication of an irrational agent. There are situations where agents may follow a strategy in which the action with highest known utility is not chosen. One such case is the maximin strategy \citep{thomson_1979}: taking the most unfavourable action that could be taken by the opponent, the maximin strategy determines how to maximise one's own utility in response to this action. Essentially, instead of maximising utility, this strategy seeks to minimise loss. Increasing attention has also been placed on the distributional reinforcement learning framework \shortcites{bellemare_2023}\citep{bellemare_2023}, which takes into account the full distribution of the reward or return and attempts to minimise loss in this way. Some approaches have also combined the minimax algorithm and distributional learning in order to improve performance \shortcites{ren_2020,li_2023}\citep{ren_2020,li_2023}. While this behaviour appears irrational, cases have been studied where playing ‘rationally’ (according to Nash Equilibria) can be shown to result in suboptimal outcomes. Notably, when playing against irrational agents, or agents that are following strategies from a different Nash Equilibrium, employing the ‘optimal’ strategy may result in lower payoff \citep{ganzfried_2023}. Employing a loss-minimising strategy may also be advantageous in adversarial settings; however, purely conflicting-interest scenarios are rare in the real world, instead we are generally operating in mixed-motive or common-interest environments \citep{dafoe_2020}.

Departing from the maximin strategy, \citet{ganzfried_2023} proposes a \textit{safe equilibrium} that balances between the maximin and Nash Equilibrium strategies by calculating the probability that the opponent is acting rationally. It is interesting to note that Ganzfried does not study this type of behaviour as a type of irrational behaviour, but as a strategy to use in interactions with potentially irrational agents (this will be discussed in more depth in the following sections). Other types of profit non-maximising behaviour can in a similar way be viewed as a strategy to minimise loss when faced with a potentially irrational opponent rather than as an irrational behaviour in itself. 

In contrast, \shortcites{goto_2011}\citet{goto_2011} present a similar method, however they do denote this behaviour as \textit{partially irrational}. In their approach, artificial learning agents take decisions with low utility with a specified probability – their rationale for these ‘irrational’ actions is to give the agent a possibility to escape from suboptimal Nash Equilibria. The approach they present can be closely assimilated to exploration within reinforcement learning, and similarly is shown to sometimes arrive at a more optimal solution than a strategy following perfect rationality would. Their model also exemplifies how game theoretic concepts can be integrated into learning algorithms to produce agents that learn to make decisions in multi-agent settings.

\subsection{Human Irrationality in Artificial Agents}
Among the types of irrationality we consider, artificial agents that contain or model human irrationality have received most attention \shortcites{armstrong_2018,uprety_2018,chen_2020,chan_2021,chen_chang_howes_2021,mcelfresh_2021,stella_2023,skalse_2023,ghosal_2023}\citep[e.g.,][]{armstrong_2018,uprety_2018,chen_2020,chan_2021,chen_chang_howes_2021,mcelfresh_2021,stella_2023,skalse_2023,ghosal_2023}. Two categories can be identified among agents of this class: the first are agents where human irrationality or cognitive biases have been intentionally incorporated, and the second is ones where human biases arise unintentionally, generally due to the data inputs - this distinction will be discussed in more depth in Section \ref{sec5}. The former has been shown to be a useful way to understand human mechanisms and learn how to improve interactions between humans and artificial agents \citep{uprety_2018}. 

Whereas reinforcement learning was inspired by the way humans and other animals learn and make decisions under uncertainty \citep{littman_2015}, inverse reinforcement learning (IRL) methods have been used to try and infer a reward function from observed behaviour. \citet{skalse_2023} look at the most common IRL methods, namely optimality \citep{ng_russell_2000}, Boltzmann rationality \citep{ramachandran_amir_2007}, and causal entropy maximisation \citep{ziebart_2010}, and test how robust these models are to misspecification. They find the most robust to be the Boltzmann-rational model, an important finding given that IRL models of human behaviour will always be misspecified to some degree. Boltzmann rationality \shortcites{luce_1959,ziebart_bagnell_2010}\citep{luce_1959,ziebart_bagnell_2010} is defined as that which ``predicts that a human
will act out a trajectory with probability proportional to the exponentiated return they receive for
the trajectory" \shortcites{laidlaw_2022}\citep[p.1]{laidlaw_2022}. However, others have argued that IRL models that assume noisy rationality, including Boltzmann rationality, are not suitable as the way humans deviate from rational behaviour is not merely noisy, but is instead systematic \shortcites{evans_2015_a,evans_2015_b}\citep{evans_2015_a,evans_2015_b}. \citet{armstrong_2018} find noisy rationality to be too strong of an assumption as it fails to account for bias, whereas a weaker simplicity assumption is also insufficient, and suggest more work is needed between the two extremes. \citet{chan_2021} similarly find that models of human behaviour as noisy-rational rather than systematically irrational perform significantly worse than those incorporating cognitive biases; in this case, Boltzmann rationality is used to model a noisy-rational agent, which is compared to systematic deviations from the Bellman equation to model irrationality. They further show that correctly modelling an agent’s irrationality can lead to higher performance than inference from rational ones.  

Others have shown that when artificial agents adopt cognitive biases observed in humans, there are cases where the performance of the agents can be improved, particularly when interacting with humans. \citet{chen_2020} demonstrate that introducing option comparison when making decisions under uncertainty can outperform calculation-based methods in highly noisy environments. While option comparison is often regarded as an irrational behaviour, they show that this heuristic is attained by a learning agent that is maximising cumulative reward. In the realm of language learning, models that acquire human biases can also be useful in better understanding existing prejudices and stereotypes \shortcites{caliskan_2017}\citep{caliskan_2017}.

\begin{table}
\caption{Types of irrationality that may result in optimal outcomes.}\label{types}
\begin{tabularx}{\textwidth}{L{0.3} L{0.3} L{0.4}} 
\toprule
 \textbf{Type of Irrationality} & \textbf{When is it Optimal?} & \textbf{Literature} \\ 
 \midrule
 \multirow{7}{=}{Bounded rationality} & \multirow{7}{=}{Limited resources} &  \citet{simon_1957,simon_1982}  \\
                                         & & \citet{russell_1997} \\
                                         & & \citet{lee_2021} \\
                                         & & \citet{simsek_2020} \\
                                         & & \citet{wen_2020} \\
                                         & & \citet{gigerenzer_2001} \\
                                        & & \citet{hüllermeier_2021} \\  
 \midrule
 \multirow{9}{=}{Random behaviour} & \multirow{9}{=}{RL exploration\\ Mixed Nash Equilibria} &  \citet{shirado_2017} \\
                         & & \citet{icard_2021} \\ 
                         & & \citet{ladosz_2022} \\
                         & & \citet{still_precup_2012} \\
                         & & \citet{fernandez_2006} \\
                         & & \citet{mnih_2015} \\
                         & & \citet{polvara_2018} \\
                         & & \citet{yu_2020} \\
                         & & \citet{kobayashi_2007} \\                        
 \midrule
 \multirow{5}{=}{Profit non-maximising} & \multirow{5}{=}{Irrational opponents\\ Adversarial scenarios}  & \citet{thomson_1979} \\
                                        & & \citet{li_2023} \\
                                        & & \citet{ren_2020} \\
                                        & & \citet{ganzfried_2023} \\   
                                        & &  \citet{goto_2011}\\                                        
\midrule
 \multirow{10}{=}{Human irrationality} & \multirow{10}{=}{Human-AI interaction} &  \citet{armstrong_2018} \\
                                                                              & & \citet{chan_2021} \\ 
                                                                              & & \citet{chen_2020,chen_chang_howes_2021} \\
                                                                              & & \citet{ghosal_2023} \\
                                                                              & & \citet{mcelfresh_2021} \\
                                                                              & & \citet{skalse_2023} \\
                                                                              & & \citet{stella_2023} \\
                                                                              & & \citet{uprety_2018} \\ 
                                                                              & & \citet{evans_2015_a,evans_2015_b} \\
                                                                              & & \citet{caliskan_2017} \\

\botrule
\end{tabularx}
\end{table}

\subsection{Perceived Irrationality in GenAI}\label{sec3.5}

GenAI systems present a slightly different case compared to the types of irrationality outlined above. Instead of having a case where irrational behaviour can in fact be optimal, we have multiple ways of assessing the rationality of a system. There is a tension between the \textit{generative rationality} and rationality that we perceive or ascribe to the system. Taking the example of LLMs, generative rationality is achieved by minimising loss in next-token prediction, whereas we interpret the output using a more human lens. As we have seen, because the output of GenAI systems such as LLMs is in a form that humans ascribe meaning to, we therefore evaluate the output not in terms of how well it achieved the goal set in the model's architecture; instead, we often evaluate these models using tasks designed to evaluate human reasoning \citep{binz_schulz_2023,macmillan-scott_2023,hagendorff_2023}. We may look for characteristics like coherence or accuracy in LLM outputs. If, instead of language, the model simply produced the embeddings as output, we would likely evaluate them as (boundedly) rational. Therefore, these models often appear irrational not because they are incorrectly minimising loss, but because there is a \textit{rationality entanglement}: overlapping types of rationality that are sometimes at odds. In fact, these hallucinations, contradictions, or other observed behaviours may be rational under the model's objective function. 

The issue of rationality entanglement arises due to our way of interpreting the model's output. We could refer to this as \textit{interpreted} or \textit{perceived} irrationality rather than actual irrationality, as we are in a sense evaluating these models incorrectly. The issue of interpreted irrationality can also be seen in other multi-agent settings that do not involve GenAI, where behaviour is merely interpreted as irrational, even though an agent is correctly maximising expected utility. This interpretation may arise for a variety of reasons, such as an incorrect assumption about the opponent's reward function, or noisy information about an opponent's actions. 

The behaviour that we interpret as irrational in GenAI systems therefore arises because we are not evaluating how well the model achieves the goal it was designed to attain. There is a misalignment in the objectives rather than a failure in reasoning. Using a traditional definition, an LLM would display rational reasoning by predicting the most likely next token, but instead we want it to produce true and consistent texts. As we will further explore below, additional layers like alignment and safety fine-tuning often in fact reduce the rationality of a model in the sense of loss minimisation. These additional training layers generally attempt to ensure that these models produce outputs that conform more closely to human expectations of rationality, even if this is at the expense of the rationality of the model in a more traditional sense. Are we then training GenAI models to behave as rational agents, or merely to behave in ways that appear normatively rational to humans?

\section{Dealing with Irrational Artificial Agents}\label{sec4}
\subsection{Identifying Irrational Artificial Agents}
We have seen that different types of seemingly irrational agents may in fact be pursuing the optimal policy when operating in the right environment. However, this creates complications for agents modelling other agents, as the typical assumption of other agents’ rationality may lead to a suboptimal outcome. \citet{albrecht_stone_2018} note that accurate modelling of other agents is particularly important in the absence of coordination and communication protocols. The first step is therefore to identify whether an agent is rational or not, and if faced with an an irrational agent, how to identify the type of irrational behaviour. 

Methods for identification of agent rationality fall under the category of opponent modelling (for surveys, see \shortcites{furnkranz_2001,herik_2005,olorunleke_2005,albrecht_stone_2018,nashed_2022}\citet{furnkranz_2001,herik_2005,olorunleke_2005,albrecht_stone_2018,nashed_2022}), an early example of which is fictitious play \citep{brown_1951}. These techniques often need to make assumptions not only about the modelled agent itself, but also about the environment – in particular relating to the observability of other agents’ action \citep{albrecht_stone_2018}. Building an accurate model of observability of the environment is especially relevant when interacting with potentially irrational agents, as we have seen that modelling irrational behaviour as simply noisy can lead to much worse performance \citep{armstrong_2018,chan_2021}. The work on \textit{rational verification} is also relevant here, particularly in its application to multi-agent systems \shortcites{abate_2021,hammond_2021}\citep{abate_2021,hammond_2021}. Rational verification checks whether given temporal logic properties will hold in a system, assuming that all agents within this system will act rationally and so choose strategies that form a game-theoretic equilibrium \shortcites{gutierrez_2021}\citep{gutierrez_2021}. In building these methods, a distinction must also be made between irrationality and unreliability – whereas in the former, observed behaviours cannot be ignored, the latter relates to noisy observations that did not in fact happen and as such should not be used in modelling an agent’s intent or reward function \citep{masters_sardina_2021}.

Several opponent modelling techniques exist in the literature, including goal/plan recognition \shortcites{ramirez_geffner_2011,tian_2016,vered_kaminka_2017,masters_sardina_2021}\citep{ramirez_geffner_2011,tian_2016,vered_kaminka_2017,masters_sardina_2021}, policy reconstruction \citep{ganzfried_sandholm_2011,chakraborty_stone_2014,silver_2016,mealing_shapiro_2017}, recursive reasoning \shortcites{hoek_wooldridge_2002,sonu_doshi_2015,weerd_2017,wen_2020}\citep{hoek_wooldridge_2002,sonu_doshi_2015,weerd_2017,wen_2020} and type-based reasoning \shortcites{he_2016,albrecht_stone_2017}\citep{he_2016,albrecht_stone_2017} – for a comprehensive survey of these methodologies, see \citep{albrecht_stone_2018}. Some of these methods directly assume rationality of a certain type: recursive reasoning techniques, such as \citeauthor{wen_2020}'s \citeyearpar{wen_2020} discussed above, assume the agent being modelled is boundedly rational, and reasons with a given level of recursion. Classification instead attempts to correctly predict a label or category for the opponent \citep{albrecht_stone_2018} – although in this case the way the modelled agent deviates from perfect rationality is not assumed from the outset, the observed behaviour is classified within existing models. In contrast, some goal recognition techniques have been proposed that calculate a rationality measure based on the agent’s action history, and use this to inform the formula used to model the agent \citep{masters_sardina_2021}. 

Work on agent modelling techniques has seen many recent developments \shortcites{zhang_2021}\citep{zhang_2021}. However, numerous open problems and questions remain, in particular regarding the rationality of agents in an interaction. Knowing in what scenarios we expect to encounter different types of irrational agents would allow for more efficient and targeted techniques. A reinforcement learning agent is likely to exhibit random behaviour as part of exploration, and it would be a mistake to interpret this as irrational; in contrast, when interacting with a human agent, observed random behaviour may be a more accurate signal of irrationality. Similarly, the type of agent that is interpreting the behaviour is important: LLMs have been shown to display behaviour that appears irrational to humans \citep{macmillan-scott_2023}, but its output is correct according to the specifications of the model, and therefore is rational from a technical perspective. Therefore, developments in this area require domain-specific research. Other open questions pertain to the possible combinations of opponent modelling techniques, as well as safe probing and exploration techniques \citep{albrecht_stone_2018}.

Safe exploration is of particular importance as agents that appear irrational may in fact be rational but deceptive agents. It may be very difficult to distinguish between the two, and particularly costly if the opponent turns out to be deceptive. Existing methods for dealing with deceptive agents remain largely domain-specific \citep{masters_sardina_2021}; some have looked at ways to reveal an agent’s true intent \citep{tambe_rosenbloom_1995}, others have focused on disambiguation of goals \citep{mao_gratch_2004,sukthankar_sycara_2005}, or have focused on Stackelberg security games where there is a potential for the attacker to be untruthful \shortcites{gan_2019}\citep{gan_2019}  – for a survey, see \citet{avrahami-zilberbrand_2014}. A clear application of safe exploration is to problems in cyber-security, where deception plays a central role \shortcites{pawlick_2019}\citep{pawlick_2019}, as in this area a priority may be mitigating the potential harmful effects of a deceptive agent. 

\begin{table}
\caption{Identifying irrational artificial agents summary table.}\label{identification}
\begin{tabularx}{\textwidth}{L{0.5} L{0.5}} 
\toprule
 \textbf{Topic}  & \textbf{Literature} \\
 \midrule
 \multirow{5}{=}{Opponent modelling surveys} &  \citet{furnkranz_2001}  \\
                                         &  \citet{herik_2005} \\
                                        &  \citet{olorunleke_2005} \\
                                        &  \citet{albrecht_stone_2018} \\ 
                                        &  \citet{nashed_2022} \\ 
 \midrule
 Fictitious play & \citet{brown_1951} \\
 \midrule
\multirow{2}{=}{Inferring reward function considering irrationality} &  \citet{armstrong_2018}  \\
                                                                     &  \citet{chan_2021} \\
 \midrule
 \multirow{3}{=}{Rational verification} &  \citet{abate_2021} \\
                                        &  \citet{hammond_2021} \\
                                        &  \citet{gutierrez_2021} \\  
\midrule
 \multirow{4}{=}{Goal/plan recognition } &  \citet{ramirez_geffner_2011} \\
                                        &  \citet{tian_2016} \\
                                        &  \citet{vered_kaminka_2017} \\  
                                        &  \citet{masters_sardina_2021} \\  
\midrule
 \multirow{4}{=}{Policy reconstruction} &  \citet{ganzfried_sandholm_2011} \\
                                        &  \citet{chakraborty_stone_2014} \\
                                        &  \citet{silver_2016} \\  
                                        &  \citet{mealing_shapiro_2017} \\  
\midrule
 \multirow{4}{=}{Recursive reasoning} &  \citet{hoek_wooldridge_2002} \\
                                        &  \citet{sonu_doshi_2015} \\
                                        &  \citet{weerd_2017} \\  
                                        &  \citet{wen_2020} \\  
 \midrule
\multirow{2}{=}{Type-based reasoning} &  \citet{he_2016}  \\
                                     &  \citet{albrecht_stone_2017} \\
\midrule
Multi-agent reinforcement learning & \citet{zhang_2021} \\
\midrule
\multirow{3}{=}{Assessing rationality in LLMs} & \citet{binz_schulz_2023} \\
                                                & \citet{hagendorff_2023} \\
                                                & \citet{macmillan-scott_2023} \\
\midrule
Safe exploration survey  & \citet{avrahami-zilberbrand_2014} \\
\midrule
 \multirow{6}{=}{Safe exploration} &  \citet{masters_sardina_2021} \\
                                        &  \citet{tambe_rosenbloom_1995} \\
                                        &  \citet{mao_gratch_2004} \\  
                                        &  \citet{sukthankar_sycara_2005} \\  
                                        &  \citet{gan_2019} \\  
                                        &  \citet{pawlick_2019} \\  
\botrule
\end{tabularx}
\end{table}

\subsection{Interacting with Irrational Artificial Agents}
Having established that an interaction is with an irrational agent, the question then arises of determining the best strategy to maximise payoff in such a scenario. Research in this area has centred on two aspects: interacting with irrational agents in adversarial scenarios, and interactions with irrational human agents. How the presence of an irrational agent may hinder cooperation/coordination and the best strategies to achieve cooperation with such an agent has received less attention, and is an important avenue of research. Applications pertain not only to scenarios that involve artificial agents, but also within human-machine interactions. As we have seen, humans often reason and act in ways that deviate from perfect rationality, therefore machines must be able to account for this. Not only should machines be able to account for irrational human decision-making; we also need to gain a better understanding of how interactions with humans may alter the behaviour of machines \shortcites{rahwan_2019}\citep{rahwan_2019}. It may be the case that techniques used in adversarial scenarios can be adapted to non-adversarial interactions, such that we interpret an irrational agent as a type of opponent.  

In adversarial interactions, techniques have emerged to minimise an agent’s own loss, as well as techniques that attempt to shape the behaviour of the opponent(s). A notable example of the former approach is the minimax algorithm (discussed above), which establishes the best strategy to minimise loss \citep{osborne_2004}. Others have built on this idea, and developed further algorithms to adapt the minimax theorem to situations involving learning agents \shortcites{li_2019}\citep{li_2019}. Distributional reinforcement learning has also emerged as a technique that considers not only the optimal action, but instead takes into account the full distribution of the potential return \shortcites{bellemare_2023}\citep{bellemare_2023}. Whereas these approaches focus on minimising loss, \citeauthor{ganzfried_2023}'s \citeyearpar{ganzfried_2023} approach discussed above balances between the maximin algorithm and Nash Equilibrium strategies, resulting in a ‘safe equilibrium.’ The trembling hand perfect equilibrium  \citep{selten_1975} similarly attempts to account for some degree of uncertainty, although for very small values of $\varepsilon$, where $\varepsilon$ is the probability of an agent ‘making a mistake' and not following the optimal strategy. A more efficient version of the minimax approach is given by the alpha-beta pruning search algorithm, which eliminates provably irrelevant subtrees to arrive at the same optimal move as minimax \citep{Knuth_1975}. \citet{bowling_veloso_2002} propose an algorithm that builds on their Win or Learn Fast (WoLF) principle: whereas the safe equilibrium approach adapts to a measure of the opponent’s rationality \citep{ganzfried_2023}, the WoLF principle reacts to their perceived position in relation to other agents by increasing learning rate when losing, and erring on the side of caution when winning. They note that previous multi-agent learning algorithms offered either convergence or rationality, whereas the WoLF principle achieves both properties. These varying approaches propose ways to adapt to the opponent’s behaviour in such a way as to maximise reward in highly uncertain scenarios.

Another recent line of research is that of opponent shaping techniques. These techniques attempt to leverage the learning of opponents to one’s advantage. Examples of these methods include the Learning with Opponent-Learning Awareness (LOLA) algorithm \citep{foerster_2018}, the Stable Opponent Shaping (SOS) algorithm \citep{letcher_2019}, or the Meta Multi-Agent Policy Gradient (Meta-MAPG) theorem \citep{kim_2021}. \citet{lu_2022} argue that these previous methods are myopic, asymmetric and require the use of higher-order derivatives; they instead propose Model-Free Opponent Shaping (M-FOS), which formulates the problem as a meta-game and does not require a model of the opponent's underlying learning algorithm. However, the application of these models remains relatively limited, focusing on social dilemma settings like the Iterated Prisoner’s Dilemma. Further research is needed into the wider application of these methods, and the potential to combine them with loss-minimising techniques.

Aside from approaches intended for adversarial scenarios, methods have emerged that address the interaction of artificial agents with irrational humans. As mentioned in the previous section, some of these methods first attempt to model the biases and irrational behaviour exhibited by human agents \citep{uprety_2018,mcelfresh_2021}, in particular as attempting to model human behaviour as noisy-rational has been shown to lead to lower performance \shortcites{kwon_2020,chan_2021}\citep{kwon_2020,chan_2021}. \citet{azaria_2022} highlights that existing models generally assume that interactions between artificial and human agents can be modelled as purely zero-sum or fully cooperative, as such ignoring the human’s goals and potential for adaptation to the behaviour of the artificial agent. Future work can build on techniques covered in the previous section for identifying and modelling systematic irrational behaviour in human agents, and develop methods that optimise the interaction between artificial and human agents in scenarios that are closer to the real world. 

\begin{table}[h!]
\centering
\begin{tabularx}{\textwidth}{L{0.45} L{0.27} L{0.3}}
\toprule
\textbf{Problem / question} & \textbf{Algorithm / method}  & \textbf{Literature} \\
\midrule
Behaviour of intelligent machines and their interaction with humans & -  & \citet{rahwan_2019} \\
\midrule
 Minimisation of loss & Minimax & \citet{osborne_2004} \\
\midrule
 Minimax extension for robust multi-agent RL & M3DDPG & \citet{li_2019} \\
\midrule
 Consider return distribution rather than expected return & Distributional RL & \citet{bellemare_2023} \\
\midrule
 Interaction with potentially irrational opponent & Safe equilibrium & \citet{ganzfried_2023} \\
\midrule
Accounting for errors in rational decision-making & Trembling hand perfect equilibrium & \citet{selten_1975} \\
\midrule
 Optimisation of minimax & Alpha-beta pruning search & \citet{Knuth_1975} \\
\midrule
 Variable learning rate, ensuring convergence & WoLF & \citet{bowling_veloso_2002} \\
\midrule
\multirow{4}{*}{Opponent shaping} & LOLA & \citet{foerster_2018} \\
 & SOS & \citet{letcher_2019} \\
 & Meta-MAPG & \citet{kim_2021} \\
 & M-FOS & \citet{lu_2022} \\
 \midrule

\multirow{5}{5cm}{Modelling (irrational) human behaviour} & \multirow{5}{*}{-} & \citet{uprety_2018} \\
                                                                            && \citet{mcelfresh_2021} \\
                                                                            && \citet{kwon_2020} \\
                                                                            && \citet{chan_2021} \\
                                                                            && \citet{azaria_2022} \\
\botrule

\end{tabularx}
\caption{Interacting with irrational artificial agents summary table.}
\label{interaction}
\end{table}

\section{Human and AI Irrationality}\label{sec5}
\subsection{Incorporating Human Irrationality into AI}
Although we have seen that humans often do not act fully rationally, and human reasoning exhibits many cognitive biases, there are cases where we may want to incorporate aspects of human reasoning into artificial agents. Initially it may appear counterintuitive – if we have established that these biases often violate the laws of logic or probability, why would we want machines to do the same? In answering this question, it is useful to reflect on why these biases appear in human reasoning. As we have seen, psychologists argue that in certain domains, heuristics and biases found in human reasoning can be remarkably effective \citep{kahneman_tversky_1982,gigerenzer_goldstein_1996,gigerenzer_brighton_2009}. For instance, sometimes we fail to consider all available information, opting instead for quicker decision-making – in many situations, this may be a desirable strategy for humans to adopt \citep{sutherland_1992}. Similarly, we may sometimes want an artificial agent to prioritise a quicker decision as opposed to a more ‘rational’ one, in a sense following \citeauthor{simon_1957}'s \citeyearpar{simon_1957} idea of \textit{satisficing}, and \citeauthor{russell_1997}'s \citeyearpar{russell_1997,russell_2016} argument that in many cases bounded rationality is a more desirable goal than perfect rationality. It is also important to note that this is nothing new, humans have long inspired the development of AI, with notable examples including neural networks \citep{gurney_2018} and reinforcement learning \citep{littman_2015}.  

A distinction must be made between artificial agents that inadvertently incorporate human biases (primarily due to the data they have been trained on), and agents that intentionally incorporate some of these biases and heuristics. There are many examples of the former in statistical learning, such as in computer vision \citep{oneil_2016,buolamwini_2018,cook_2019,wilson_2019,noiret_2021,papakyriakopoulos_2023} and NLP \citep{bolukbasi_2016,rozado_2020,dawkins_2021,hovy_2021,stanczak_2021}, where the data used to train models contained biases and has resulted in negative or harmful consequences. Risks arise when machines that learn from human data miscategorise human irrationalities as human values and optimise for these irrationalities \citep{gorman_armstrong_2022}. With the advent of LLMs, we are now seeing agents that not only inadvertently incorporate human biases, but also mimic limitations in human reasoning and exhibit the same cognitive biases \shortcites{bubeck_2023,macmillan-scott_2023}\citep{bubeck_2023,macmillan-scott_2023,binz_schulz_2023,hagendorff_2023}. 

However, in this section we focus on the second case – on those instances in which it may be beneficial to purposefully integrate heuristics into the way artificial agents make decisions. Often, the benefits of incorporating heuristics arise in cases where there are time constraints, limited computational resources, or partial information \citep{simsek_2020}. \citet{gigerenzer_2001} presents an illustrative example of a robot attempting to catch a ball: following a rational approach, the robot would need information on all the possible trajectories the ball might follow, as well as ways to measure it's velocity and angle, among other factors. However, professional athletes use the simple heuristic of keeping their eyes on the ball as they run towards it. Using this heuristic, a robot would only need to know the angle of gaze to achieve a high performance - a much simpler, `less rational' method that achieves the same result.

\citet{gulati_2022} propose a taxonomy of cognitive biases present in human decision-making that may be valuable in the development of AI systems. While they highlight that this is particularly important for the future of human-machine collaboration, we argue that the benefits of research in this area will also be present in interactions between artificial agents, as well as in single agent settings. One example of this are the symmetry and mutual exclusion biases, which have been incorporated into a Naïve Bayes classifier \shortcites{taniguchi_2017}\citep{taniguchi_2017} and neural networks \shortcites{taniguchi_2019,manome_2021}\citep{taniguchi_2019,manome_2021}. The authors show that these methods can be used to produce models that learn from small and biased datasets, and that their models outperform more traditional machine learning methods. Others have also demonstrated how decision heuristics can outperform more complex statistical methods when there is limited data and training examples \citep{brighton_2006,katsikopoulos_2011,simsek_buckmann_2015,buckmann_simsek_2017}. Similarly, within natural language, this type of approach can remove the need for a human trainer and lead to agents that learn from their own mistakes and correct this in planning sequences \shortcites{shinn_2023}\citep{shinn_2023}. The effective use of human cognitive biases is also evident under uncertainty: \citet{chen_2020} show that an agent that integrates calculative and comparative decision-making outperforms an agent that merely calculates the optimal choice. 

The benefits of incorporating cognitive biases into artificial agents do not only relate to improving the capabilities of boundedly rational agents. The use of heuristics can also be used to enhance the explainability of AI processes \citep{newell_shaw_simon_1959,simsek_2020}, as well as to design agents that are more robust to interactions with non-rational or boundedly rational agents \citep{wen_2020}. Examples of the latter can be found in the study of population dynamics \shortcites{yang_2018}\citep{yang_2018} and video game design \shortcites{hunicke_2005,peng_2017}\citep{hunicke_2005,peng_2017}. Much of the research in this area is recent, so questions remain as to the potential benefits that can arise from incorporating heuristics derived from human decision-making into artificial agents; the application to boundedly rational agents is clearer, but it has been shown that the impact of such heuristics can extend beyond bounded rationality. Notably, research has shown that behaviours that are often thought to be irrational in humans can emerge in artificial agents from a learning process that seeks to maximise cumulative reward \shortcites{howes_2016,chen_2020}\citep{howes_2016,chen_2020}.

\subsection{Human-AI Interaction with Irrational Machines}\label{sec5.2}

Artificial agents are becoming more and more integrated into our lives, and the frequency of human-AI interactions is quickly rising \shortcites{amershi_2019}\citep{amershi_2019}. Therefore, it is important to understand how the rationality of these artificial agents impact the interaction. At the same time, it is worth noting that this relationship is bidirectional: the behaviour of machines and humans will each impact the other, and it is this co-behaviour that required a deep understanding \citep{rahwan_2019}. Small errors in the algorithm or data may have large unpredictable effects on society, and could even have the potential to alter the social fabric \citep{rahwan_2019}. As underlined by the field of AI safety, flaws in the design of AI systems may result in emergent behaviours that are potentially harmful and may pose a risk to society \shortcites{amodei_2016,hendrycks_2022}\citep{amodei_2016,hendrycks_2022}. 

Although the goal is often to create agents that are as close to perfectly rational as possible, humans may feel more comfortable in interactions with agents that are slightly irrational or unpredictable. This is particularly relevant with the widespread use of LLMs like ChatGPT. On the other hand, artificial agents that are less rational may be perceived as untrustworthy or incompetent. The discrepancy between an individual’s expectations and their perception of a smart system’s actions can lead to the human perceiving the system as irrational \shortcites{abadie_2019}\citep{abadie_2019}. The threshold for losing trust in algorithms appears to be much lower than for humans - a phenomenon referred to as \textit{algorithmic aversion} \shortcites{dietvorst_2015}\citep{dietvorst_2015}. Studies have looked at humans' perception of machines as an aid to decision-making \shortcites{binns_2018}\citep{binns_2018}, and the use of machines has been shown in cases to amplify human biases \citep{glickman_2022}. 

Some have argued that human-machine collaboration may require AI systems that replicate human cognitive biases \citep{gulati_2022}. However, it has also been shown that we do not necessarily perceive and evaluate machines in the same way as humans and other components of our environment \citep{hidalgo_2021}. Therefore, some of the biases that have been studied in human interaction may not be present in human-AI interaction, such as the social desirability bias \citep{gulati_2022}. Studies have looked at how the incorporation of human physical attributes affect the dynamics of the interaction: \citet{pizzi_2020} show how anthropomorphic agents can have a positive impact on the human’s perception of the artificial agent, increasing satisfaction and acceptance; a similar effect is found for increasing trust in agents with more anthropomorphic features \shortcites{waytz_2014}\citep{waytz_2014}. Whether this is also the case with cognitive attributes is currently not well understood \citep{shanahan_2024}. Another important aspect is human's perception of machine morality; \citet{benn_grastien_2021} propose an approach of conveying ethical understanding in human-robot interactions, highlighting that what matters is not only the morality of an agent, but also the appearance of ethical understanding indicated by its behaviour. 

\section{The GenAI Evaluation Paradox}\label{sec6}

GenAI is unique in that it generates content in a format that humans ascribe meaning to. This characteristic of generative models leads to the \textit{GenAI Evaluation Paradox}. There is a tension between what models are being trained to maximise and what they are being evaluated on \citep{goldstein_2024}. This issue can be viewed as a different kind of  performance vs. competence distinction \citep{firestone_2020}. As LLMs are the most commonly used types of GenAI systems, we can take these as an example, although the same argument applies to other GenAI systems that produce different types of output like visual or audio output (or combinations of modalities). At their core, LLMs aim to achieve predictive or generative rationality: this can be measured by looking at how good the models are at predicting the most likely next token. Another way to assess LLMs is in accordance to their epistemic rationality: the definitions of rational reasoning in philosophy and psychology discussed above that is used to evaluate human reasoning. For instance, \citet{jiang_2025} determine four necessary axioms for a rational agent: information grounding, logical consistency, invariance from irrelevant information, and orderability of preference. They clearly state that these axioms have been derived from work in cognitive psychology, so are similar to principles we would use to evaluate rational reasoning in humans.

The evaluation of LLM outputs through epistemic rationality leads to anthropomorphisation of LLMs and often results in researchers concluding that these models reason in irrational ways, such as by displaying human biases or outputting hallucinations. There is a misalignment between what we are training the models to maximise: generative rationality, and the behaviour we in actuality want these models to exhibit: epistemic rationality. There may then be a need to reevaluate how we are setting the training goals of GenAI systems to ensure that these are more in line with what we want to assess them on, whether this be accuracy, morality, or another metric. 

The behaviour exhibited by LLMs that is interpreted as irrational sometimes mirrors biases in human reasoning, whereas other times it emerges in other ways. Numerous studies have applied cognitive psychology tools to evaluate whether LLMs replicate human cognitive biases and heuristics \citep{binz_schulz_2023,macmillan-scott_2023,hagendorff_2023}, treating these models like participants in a psychology experiment and evaluating their outputs as you would for humans. The presence of this mimetic irrationality in LLM reasoning arises from the data they are trained on, which contains our own biases. However, sometimes LLMs display emergent irrational reasoning that is not present in the training data. For instance, they may hallucinate facts, contradict themselves, or give incorrect explanations. The presence of such emergent behaviour can sometimes be a product of additional training mechanisms such as Reinforcement Learning from Human Feedback (RLHF). RLHF has been shown to induce sycophantic behaviour in LLMs \citep{perez_2024}, even to the point of displaying manipulative and deceptive tendencies in order to maximise feedback from users \citep{williams_2025}. Again, this is an example where the model is performing rationally according to its goals, but not according to what we consider rational or desirable. Aside from RLHF, other types of alignment and safety fine-tuning can similarly be at odds with the model's initial objective function. These methods often alter the model's weights, and some even modify the objective function itself (e.g., RLHF or contrastive learning). These efforts generally aim to constrain the output of GenAI systems within ethical and social norms. There is also a tension between the different types of safety fine-tuning; for instance, some types of fine-tuning may attempt to improve truthfulness, whereas RLHF can counteract this by reinforcing more sycophantic tendencies, which leads to LLMs lying in order to obtain positive feedback \citep{williams_2025}. 

It is not enough to merely reevaluate how we define the model's objectives or evaluation metrics. Depending on the application and scenario that a GenAI system is being used in, we will want to maximise or curtail certain behaviours. Therefore, working to develop more specialised systems that are better suited to particular applications may mean we are better able to define what we are trying to achieve. We have already discussed that there is no unified conception of rationality within computer science, and that different types of rationality are better suited to certain applications. Rapid advancements in GenAI compel us to answer these questions with more urgency. In critical decision-making scenarios we will want to maximise properties like truthfulness and accuracy, whereas in more creative applications we may want to maximise originality. It is worth noting that virtually none of the applications that LLMs are currently applied to, if any, seek to maximise solely the performance of next-token prediction. Nonetheless, all of its current applications require the next-token prediction to work at least at a satisfactory level, leading us back to Simon's idea of \textit{satisficing} and bounded rationality \citep{simon_1957,simon_1982}. 

GenAI presents a difficult balancing act between overlapping and sometimes conflicting objectives that lead to a rationality entanglement: we need to find the right equilibrium between generative performance of next-token prediction, alignment to human preferences, epistemic coherence, and task specialisation. Disentangling the tension between these aims is crucial to inform model architectures and evaluations.

\section{Open Questions}\label{sec7}
Below we define a set of open questions in a number of areas of rationality in AI. The list is not exhaustive but sets out a series of promising research directions or areas that have much to be explored.

\subsection{Defining Rationality in AI}
\begin{enumerate}[leftmargin=1cm, itemsep=5pt]
    \item[1.] \textit{Is a general, unified definition of rationality in AI achievable?}
    \item[2.]\textit{How should rationality be defined for GenAI systems?}
    \item[3.] \textit{Should we distinguish between definitions and assessments of rationality for symbolic AI and statistical learning?}
\end{enumerate}
\vspace{0.25cm}
As we have seen, different disciplines define rationality in very different ways. The understanding of rationality within AI has been influenced by some of these, particularly by work within economics, philosophy, and psychology. As artificial agents become more complex and capable, in particular considering the emerging capabilities of foundation models, it is important to determine whether we can or should establish a unifying definition of what constitutes a rational agent. A general framework could provide common ground for evaluating reasoning and decision-making across systems, but may be too general to prove useful. It will likely not suffice to adapt a definition from human rationality; we need to develop a definition of machine rationality distinct to that of humans. Even a distinct definition for AI agents that is set apart from human rationality will likely need to be further categorised. For example, GenAI systems may need to be evaluated in light of the GenAI Evaluation Paradox. Additionally, distinctions between symbolic AI and statistical learning approaches may warrant separate classification frameworks. 

\subsection{Training and Evaluation}
\begin{enumerate}[leftmargin=1cm, itemsep=5pt]
    \item[4.] 
    \begin{enumerate}[itemsep=4pt]
        \item[(a)] \textit{What novel training or evaluation methodologies can be developed that account for the GenAI Evaluation Paradox?}
        \item[(b)] \textit{Should we develop benchmarks or testing suites that integrate multi-dimensional rationality, particularly for GenAI?}
    \end{enumerate}
    \item[5.] \textit{How can we realign the training objectives of GenAI systems with human-evaluated outputs?}
    \item[6.] \textit{Should we evaluate rational processes as well as rational output?}
\end{enumerate}
\vspace{0.25cm}
In order to avoid the apparent failures in \textit{competence} being judged because of poor \textit{performance} \citep{firestone_2020}, we need to ensure that we are training AI agents with objectives that are aligned to what they will be evaluated on. This issue is particularly pertinent for GenAI, where there is often a misalignment between training and evaluation. We are generally interested in output characteristics like accuracy or lack of harmful content, whereas, at their core, GenAI systems are not trained to maximise for these objectives. The rationality entanglement present in GenAI systems raises the question of whether we can account for rationality in multiple dimensions. Similarly, we generally focus on measuring only outputs, whereas looking at the rationality of reasoning and decision-making processes could provide novel insights. 

\subsection{Conflicts and Trade-offs}
\begin{enumerate}[leftmargin=1cm, itemsep=5pt]
    \item[7.] \textit{What are the computational and ethical trade-offs between pursuing perfect vs. bounded rationality in AI?}
    \item[8.] \textit{What are the safety implications of building systems that deviate from perfect rationality?}
    \item[9.] \textit{Can a single system maximise multiple types of rationality at once, or will there inevitably be a trade-off?}
\end{enumerate}
\vspace{0.25cm}
When setting our objectives in the design of AI agents, there are a number of conflicts and trade-offs that must be considered. Perfect rationality has been shown to be an unrealistic goal \citep{russell_1997,russell_2016,lee_2021}; in choosing to aim instead for goals like bounded rationality, we need to understand what the implications of this choice are. Systems that deviate from perfect rationality may introduce unpredictability. Also, as we saw with GenAI, depending on how we evaluate a system, there may be more than one type of rationality that can be contained and assessed within one system. In such a case, we will likely need to prioritise which type of rationality we are most interested in. 

\subsection{Understanding Irrationality in AI}
\begin{enumerate}[leftmargin=1cm, itemsep=5pt]
    \item[10.] \textit{Can we develop a generalisable framework to categorise types of irrationality in artificial agents?}
    \item[11.] \textit{In what scenarios can the incorporation of irrational behaviour improve capabilities relating to issues like trust, performance or cooperation?}
    \item[12.] \textit{Can the modelling of irrationality in AI provide insights into human reasoning and decision-making?}
    \item[13.] \textit{How can we distinguish between actual irrationality and deceptive behaviour in artificial agents, especially in adversarial settings?}
\end{enumerate}
\vspace{0.25cm}
We need to deepen our understanding not only of rationality, but also of the various forms that irrationality can take. We have seen that irrational behaviour does not necessarily lead to sub-optimal outcomes. A more general and comprehensive categorisation of different types of irrationality would be beneficial in clearly establishing when these types of behaviours can be useful and improve capabilities of AI systems or help us to investigate characteristics of human behaviour. A distinction must also be made between irrational behaviour and behaviour that is perceived to be irrational, but may in fact be deceptive. For instance, LLMs outputting inaccurate information may be a symptom of manipulative or deceptive behaviour \citep{williams_2025}.

\subsection{Human-AI Interaction}
\begin{enumerate}[leftmargin=1cm, itemsep=5pt]
    \item[14.] \textit{Can we integrate human heuristics into AI reasoning to improve interpretability?}
    \item[15.] \textit{Should AI aim to correct human irrationality or adapt to it in collaborative contexts?}
    \item[16.] \textit{How should AI systems balance exploration and exploitation in environments where human preferences are uncertain or evolving?}
\end{enumerate}
\vspace{0.25cm}
Heuristics and biases in human reasoning may at times lead us to incorrect conclusion, but they are useful in making decisions under uncertainty. Similar mechanisms can potentially be integrated into the decision-making of AI systems, and these may even improve interpretability \citep{newell_shaw_simon_1959,simsek_2020,simsek_2016}. 

Increasingly, AI agents will need to interact with humans, and there are many open questions around how to approach this. In some cases, like safety-critical scenarios, it may be beneficial to try to correct human irrationality, whereas in others the AI system may need to adapt to this behaviour. When artificial agents encounter human irrationality, we need to decide what the interaction should look like. In particular, learning from human behaviour may be more complex than learning from artificial agents due to our changing preferences and reward functions \shortcites{carroll_2024}\citep{carroll_2024}.

\subsection{Multi-Agent and Collective Rationality}
\begin{enumerate}[leftmargin=1cm, itemsep=5pt]
    \item[17.] \textit{What combinations of opponent modelling techniques yield the most robust results in environments with irrational agents?}
    \item[18.] \textit{Can we study collective rationality in AI systems in the same way that we do individual rationality?}
\end{enumerate}
\vspace{0.25cm}
Multi-agent settings with potentially irrational agents require robust opponent modelling techniques. It remains to be seen what methods can be adapted from adversarial scenarios, and what additional methods can be developed for these setting.

In this paper, although we have discussed multi-agent scenarios, the focus has been on individual rationality. Much of the literature, not just in computer science, has focused on individual rationality \citep{knauff_2021}. We will likely need to take a different approach to better understand collective rationality, distinguishing between types of agents, as well as between scenarios involving only artificial agents and those including humans. Collective or social rationality adds additional dimensions, like group dynamics, communication and shared goals.

\section{Conclusion}
Rationality and irrationality play an important role within AI. Designing the `rational agent' has been the objective and the focus of research and practice in the field for many years. However, achieving this requires a common understanding of what it means to be rational. As we have seen, definitions of rationality within economics, philosophy, and psychology have informed our understanding of the concept within AI. We may aim for different types of rationality, as certain types will be more suitable in some contexts. Crucially, a distinction must be made between how we define rationality in humans and machines. This article provides a survey of the existing literature and methods in this area. Having established that particular irrational behaviour may in fact be optimal under certain conditions, it is clear that the issue is complex. When it comes to GenAI, additional questions and tensions arise regarding how we train and evaluate these models. From a methodological perspective, techniques have emerged for the identification and interaction with irrational agents. However, these remain scarce – further research is needed in this area, both looking at how existing opponent modelling or shaping methods may be adapted to account for irrationality, as well as how they may be combined to better suit the problem at hand. The question of interacting with irrational agents is crucial not only among machines, but also because humans often act in irrational ways. Human-AI interaction is a key aspect of today's AI systems, namely with the case of systems based on LLMs and their widespread use. Cognitive biases may in some instances be leveraged to improve the performance of artificial agents, whereas in human-AI interaction the design of machines will need to account for the irrational behaviour of humans. At the same time, further understanding is needed around how the rationality of an artificial agent impacts the dynamics of human-AI interaction - there may be cases where it is beneficial for the artificial agent to appear not fully rational, whereas in other cases this can deem the agent untrustworthy. This article lays out open questions relating to rationality within AI. As we strive to create more capable agents that are increasingly integrated in everyday lives, these questions will need to be addressed. 

\vskip 0.5in

\section*{Declarations}

\textbf{Competing interests.} The authors have no competing interests to declare.

\vspace{0.1in}

\noindent \textbf{Funding.} No funding was received to assist with the preparation of this manuscript.

\vspace{0.1in}

\noindent \textbf{Author contributions.} M.M. and O.M.-S. conceptualised the survey. O.M.-S. wrote the manuscript text and M.M. supervised the work. M.M. and O.M.-S. reviewed the manuscript.

\vspace{0.1in}

\noindent \textbf{Acknowledgements.} The authors would like to thank the anonymous reviewers for their constructive comments and valuable suggestions, which have helped to improve the quality and scope of this manuscript.

\vskip 0.2in
\bibliography{main}


\end{document}